# Spatial-Channel Transformer Network for Trajectory Prediction on the Traffic Scenes

Jingwen Zhao, Xuanpeng Li*, *Member, IEEE*, Qifan Xue, and Weigong Zhang, *Member, IEEE*

*Abstract*—Predicting motion of surrounding agents is critical to real-world applications of tactical path planning for autonomous driving. Due to the complex temporal dependencies and social interactions of agents, on-line trajectory prediction is a challenging task. With the development of attention mechanism in recent years, transformer model has been applied in natural language sequence processing first and then image processing. In this paper, we present a Spatial-Channel Transformer Network for trajectory prediction with attention functions. Instead of RNN models, we employ transformer model to capture the spatial-temporal features of agents. A channel-wise module is inserted to measure the social interaction between agents. We find that the Spatial-Channel Transformer Network achieves promising results on real-world trajectory prediction datasets on the traffic scenes.

## I. INTRODUCTION

Autonomous vehicles need to make safe and efficient decisions to pass through the complex traffic scenes, such as changing lanes, overtaking or decelerating. It can be seen in the existing tactical path planning algorithms[1] that these algorithms require reliable estimate of the future trajectories of surrounding agents, like vehicles and pedestrians. Therefore, autonomous vehicles need to have the ability to infer the future movements of surrounding agents.

Trajectory prediction of multiple agents utilizes the history trajectories and behaviors to forecast their future motion, which is closely related to the current traffic condition. Due to the uncertainty of behavior, trajectories tend to be highly non-linear during a long time. Additionally, the potential interactions between surrounding agents affect action decisions. Therefore, it is challenging to model and predict the trajectory of multiple agents on the traffic scenes.

With the development of deep neural networks, Long Short-Term Memory (LSTM) network has become the primary method for trajectory prediction. LSTM sequentially processes time series data to characterize the speed, direction and motion mode of agents. Besides, social pooling mechanisms are designed to simulate the social interaction of agents[2, 3]. However social pooling treats agents equally through pooling operations, which weakens the impact of their interactions. Attention mechanisms[4-6] are appended to weigh surrounding agents unequally based on a learning process. Nonetheless the existing trajectory prediction approaches still have two common limitations:

- Since time dependence is complex to extract[7], LSTM has been criticized for the memory mechanism[8] and the ability to model social interaction[9].

- The utilization of attention mechanism is too simple to properly simulate the interaction between vehicles;

Recently, the Transformer model made breakthrough progress in sequence learning and sequence generation tasks in natural language processing domains[7, 10], which inspires us to replace LSTM model with Transformer to extract time dependencies better. In this paper, we propose the Spatial-Channel Transformer framework, a novel model with completely attention mechanism to deal with trajectory prediction problems on the traffic scenes. With powerful self-attention mechanism and multi-head attention mechanism, the Encoder-decoder Transformer Network models agent motions by the spatial position and temporal correlation. Additionally, we employ Squeeze-and-Excitation Blocks[11] to capture the interaction between agents with channel-wise attention. The unified experiments of multiple classic approaches are conducted on the NGSIM dataset with several metrics. We also carry out ablation studies to highlight the effect of component modules.

## II. RELATED WORK

RNNs and their variants LSTM[12] and GRU[13] have achieved great success in sequence prediction tasks such as robot localization[14], and robot decision making[15]. RNNs have also been successfully applied to simulate the motion patterns of pedestrians or vehicles with a Seq2Seq structure[16]. In addition, to improve trajectory prediction, social pooling mechanism[2, 3], attention mechanism[4, 17] and graph neural network[6, 18] are used to simulate interactive behaviors.

Social LSTM[2] is one of the first deep learning models to study pedestrian trajectory prediction. Social LSTM uses a pooling mechanism to aggregate the hidden feature output and predict the following trajectory. CS-LSTM aims at researching on vehicle trajectory prediction, which improves social pooling mechanism with convolution operation[19]. Later models State-Refinement LSTM (SR-LSTM)[5] extended Social LSTM through visual features and new pooling mechanism to improve prediction accuracy. Especially SR-

*Research supported by National Natural Science Foundation of China under Grant 61906038.

Jingwen Zhao, Xuanpeng Li, Qifan Xue and Weigong Zhang are with the School of Instrument Science and Engineering, Southeast University, Nanjing, 210096, China (corresponding author: Xuanpeng Li, phone: +8613770666889, e-mail: li_xuanpeng@seu.edu.cn).

Figure 1. Structure of the Spatial-Channel Transformer Network.

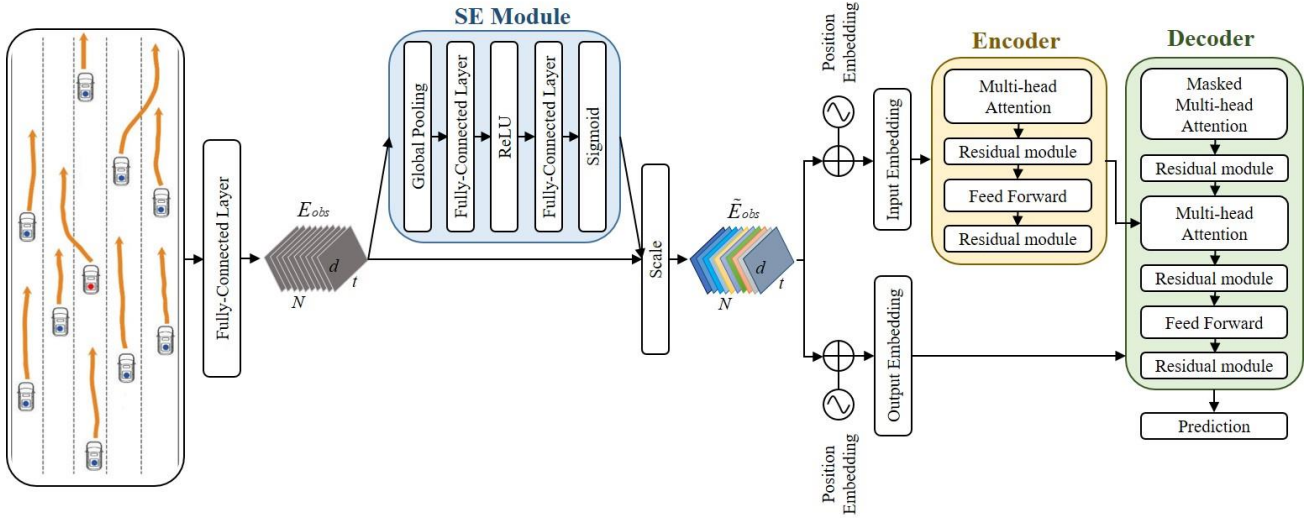

LSTM weights the contribution of each pedestrian to others through a weighting mechanism.

Considering that the structure of the graph network is suitable for modeling interaction, several approaches successfully used Graph Neural Networks (GNN) to characterize the influence among agents, including STGAT model[6] with GAT module and Social-STGCNN[18] with GCN module.

Additionally, generative model is also well applied in trajectory prediction. Based on the assumption of the multi-modal distribution of pedestrian trajectories, Social GAN[3] extended Social LSTM to a generative model. Sophie[20] uses CNNs to extract features from the entire scene with a two-way attention mechanism for each pedestrian. Sophie concatenates the attention output with the visual CNN output, and then uses a generative model based on LSTMs auto-encoder to generate future trajectories.

In recent years, the transformer network has dominated the field of Natural Language Processing[7, 10]. It completely abandons the recurrence and focuses on the attention over time step. This architecture has the ability of parallel training and capturing long-term dependency, which can still predict from inputs with missing observation data. Recently the transformer architecture has also been successfully applied in pedestrian trajectory prediction. This paper provides a transformer application of trajectory prediction on the traffic scene. Without using the graph structure, we use complete attention mechanism to process time-series data and model social interaction, which provides proof of the validity of the attention mechanism on sequence prediction.

Furthermore, Squeeze and Excitation (SE) Networks[11] bring significant improvements in performance of existing CNN models for image processing. We are inspired by the Squeeze and Excitation module and apply it to the trajectory embedding with channel-wise attention to capture interaction between neighbors.

## III. MODEL

In this section, we introduce the trajectory prediction model based on spatial-channel attention. The Encoder-decoder Transformer Network is used to extract temporal sequence features and utilize the past spatial positions of multiple agents to predict future trajectories. The SE module is added to capture the impact of agents' interaction on the past and future trajectory data.

### A. Problem Formulation

We assume that there are $N$ agents involved on a traffic scene. The input of our model consists of the trajectories of agents, represented as $X^t = [X_1^t, X_2^t, X_3^t, ..., X_N^t]$, where $X_i^t = (x_i^t, y_i^t)$ is defined by its x-y coordinates. We observe the positions of all the agents from time 1 to $T_{obs}$, and predict their positions for time $T_{obs+1}$ to $T_{pred}$. The future trajectories of ground truth can be represented as $Y^t = [Y_1^t, Y_2^t, Y_3^t, ..., Y_N^t]$, and the output sequence of our model can be denoted as $\hat{Y}^t = [\hat{Y}_1^t, \hat{Y}_2^t, \hat{Y}_3^t, ..., \hat{Y}_N^t]$.

### B. Encoder-Decoder Transformer Network

The Encoder-Decoder structure is extensively used in sequence transduction models[16]. In this work, the encoder network maps input sequence $X^t = [X_1^t, X_2^t, X_3^t, ..., X_N^t]$ to latent state $Z^t = [Z_1^t, Z_2^t, Z_3^t, ..., Z_N^t]$. With the given $Z^t$, the decoder network generates one element of the output sequence $\hat{Y}^t = [\hat{Y}_1^t, \hat{Y}_2^t, \hat{Y}_3^t, ..., \hat{Y}_N^t]$ on every time step. In an auto-regressive way, the previously generated element is used as an additional input to generate the next element.

As illustrated in Fig.1, the Encoder-Decoder Transformer Network is composed of several blocks, including Attention module, Feed-Forward Networks module, and Residual Dropout module.

The Transformer network mainly captures the dependencies of time-series data and the non-linear characteristics of spatial data through the Attention module, which comprises of self-attention mechanism and multi-head attention mechanism.

Self-attention takes query matrix $Q$, key matrix $K$, value matrix $V$, dimension $d_k$ of queries and keys as input, which is calculated by a dot product of $Q$ and $K$, and scaled factor of $1/\sqrt{d_k}$. Then a softmax function is applied to obtain the weights of values. The scaled operation is better for large value of $d_k$ to avoid the softmax function falling into regions where it has extremely small gradients.

$$\text{Attention}(Q,K,V) = \text{softmax}(\frac{QK^T}{\sqrt{d_k}})V \quad (1)$$

Instead of learning a single attention function, it has found that it is beneficial to map the queries, keys and values for h times to learn different contextual information respectively[7]. The self-attention function is performed on each projected version of queries, keys and values in parallel. Then the results are concatenated and projected again to obtain the weight of final values. Therefore, benefitting from the multi-head attention operation, the transformer structure can jointly generate comprehensive latent feature of trajectory data from different representation subspaces.

$$\text{MultiHead}(Q,K,V) = \text{Concat}(\text{head}_1, \ldots, \text{head}_h)W^O \quad (2)$$

Where $\text{head}_i = \text{Attention}(QW_i^Q, KW_i^K, VW_i^V)$. $W_i^Q$, $W_i^K$, $W_i^V$, $W_i^O$ are the projections of parameter matrices in queries, keys, values and output.

Besides, the fully connected Feed-Forward Networks consists of two linear transformations with a ReLU activation, which is applied to each attention sub-layers. And the addition of the Residual Dropout module is set to improve the efficiency of the network.

$$\text{Feedforward}(x) = \max(0, xW_1 + b_1)W_2 + b_2 \quad (3)$$

*C. Squeeze and Excitation Module*

In order to deal with social interaction between vehicles, we embed $N$ trajectories of vehicles in time into a vector $E_{obs}$ with dimension of $N \times T_{obs} \times d$, and use $N$ as the number of feature channels of the input tensor, as illustrated in Fig.1. With the SE module for channel-wise attention, the interactions between vehicles are extracted in weights. The SE module consists of squeeze operation, excitation operation, and scale operation.

The squeeze operation manipulates spatial dimensions to perform feature compression, turning each two-dimension feature channel into a real number. This real number has a global receptive field of feature representation of each vehicle trajectory, and the output dimension matches the input feature channel number $N$. It characterizes the global distribution of the feature channel, which is consistent with the interaction between adjacent vehicles.

The excitation operation is similar to the gate mechanism in the recurrent neural network. The weight is generated for each feature channel through the parameter $w$, where the parameter $w$ is learned to explicitly model the correlation between the feature channels.

The scale operation treats the results of the excitation as the importance of each feature channel, which reweighs the original input feature in the channel dimension through multiplication.

$$z_c = F_{sq}(E_{obs}) = \frac{1}{T_{obs} \times d} \sum_{i=1}^{T_{obs}} \sum_{j=1}^{d} E_{obs}(i,j)$$
$$s_c = F_{ex}(z_c, w) = \sigma(g(z_c, w)) = \sigma(w_2 \delta(w_1 z_c)) \quad (4)$$
$$\tilde{E}_{obs} = F_{scale}(E_{obs}, s_c) = s_c E_{obs}$$

In this work, global average pooling is used as for the squeeze operation. Then two fully connected layers in a bottleneck structure estimate the correlation between channels and output the same number of weights as the input feature channels, where δ refers to the ReLU function. Through a sigmoid gate σ, the normalized weights $s_c$ between 0 and 1 are obtained, which will be multiplied by the original feature to represent the social interaction between vehicles through the channel-wise attention.

*D. Implementation Details*

Different from the LSTM model, the transformer network discards the sequential nature of time-series data and models temporal dependencies with the self-attention mechanism. Therefore, the input embedding data $E_{obs}^{(i,t)}$ consists of spatial trajectory embedding and temporal position embedding. For the trajectory embedding $e_{obs}^{(i,t)}$, we expand the dimension of trajectory sequences $X^t = [(x^{(i,t)}, y^{(i,t)})]$ from 2 to $D = 512$ with MLP network. And the position embedding $P^t$ is defined by sine and cosine functions.

$$E_{obs}^{(i,t)} = e_{obs}^{(i,t)} + P^t$$
$$P^t = \{p_{(t,d)}\}_{d=1}^{D}$$
$$p_{(t,d)} = \begin{cases} \sin(t/10000^{d/D}) & \text{for } d \text{ even} \\ \cos(t/10000^{d/D}) & \text{for } d \text{ odd} \end{cases} \quad (5)$$

In this work, the Encoder-Decoder Transformer Network is composed of 8 attention heads. The loss function is defined by L2-loss between the predicted output trajectory $\hat{Y}^t$ and the ground truth trajectory $Y^t$. Via backpropagation with the Adam optimizer, we trained the network with learning rate 0.01 and dropout value of 0.1. The training and inference is implemented using PyTorch.

## IV. EXPERIMENTS

In this section, we evaluate our method on the publicly available NGSIM US-101 and I-80 datasets. Each dataset consists of 15-minute segments of mild, moderate and congested traffic conditions. In the same way of the previous work[19], we split the complete dataset into train, validation and test sets. With the local coordinate data provided by each dataset, we split the trajectories into segments of 8 seconds, where we use 3-second history observation and next 5 seconds for prediction. The original sampling frequency of the 8-second segments is 10 Hz. We subsampled it by a factor of 2 as previous work[19].

Table 1. The ADE, FDE, RMSE performance of Spatial-Channel Transformer Network are presented (in meter) over a 5-second prediction horizon compared with baseline models on NGSIM datasets.

|    |         | Social LSTM | CS-LSTM   | Social STGCNN | Social GAN | Social STGAT | Ours      |
|----|---------|-------------|-----------|---------------|------------|--------------|-----------|
| 1s | ADE/FDE | 0.29/0.53   | 0.25/0.48 | 0.50/0.80     | 0.16/0.33  | 0.15/0.33    | 0.20/0.39 |
|    | RMSE    | 0.61        | 0.57      | 0.72          | 0.30       | 0.31         | 0.38      |
| 2s | ADE/FDE | 0.67/1.46   | 0.62/1.38 | 0.92/1.71     | 0.43/0.99  | 0.43/1.00    | 0.50/1.10 |
|    | RMSE    | 1.35        | 1.28      | 1.36          | 0.78       | 0.80         | 0.88      |
| 3s | ADE/FDE | 1.19/2.78   | 1.12/2.65 | 1.40/2.82     | 0.79/1.90  | 0.79/1.88    | 0.88/2.06 |
|    | RMSE    | 2.23        | 2.14      | 2.10          | 1.37       | 1.39         | 1.51      |
| 4s | ADE/FDE | 1.84/4.51   | 1.75/4.33 | 1.95/4.16     | 1.23/3.02  | 1.22/2.94    | 1.35/3.26 |
|    | RMSE    | 3.36        | 3.23      | 2.96          | 2.09       | 2.10         | 2.27      |
| 5s | ADE/FDE | 2.62/6.62   | 2.51/6.43 | 2.57/5.71     | 1.93/4.64  | 1.89/4.40    | 1.90/4.66 |
|    | RMSE    | 4.63        | 4.59      | 3.93          | 3.17       | 3.10         | 3.16      |

Figure 2. Trajectory visualization. We list the trajectory prediction results of Social GAN, STGAT and Spatial-channel model in the same scene for comparison. The red line is represented as the historical observation for input. The ground truth trajectories are shown as points in green, and the blue points make up the results of prediction.

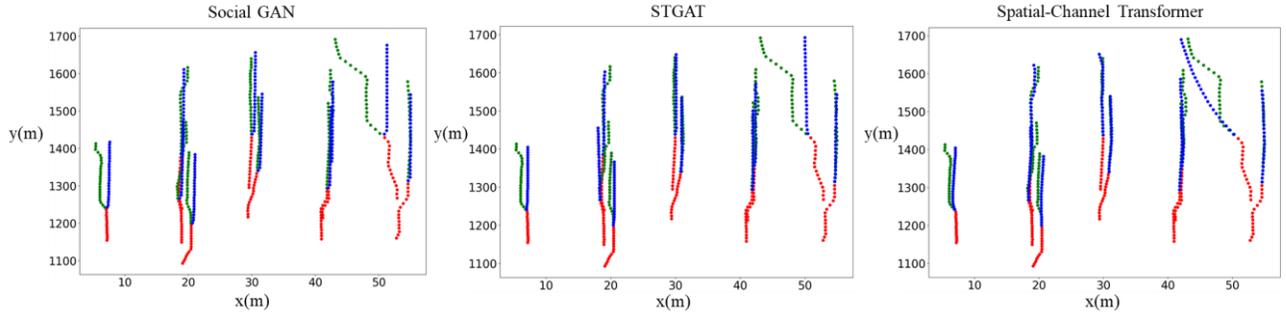

### A. Evaluation Metrics and Baselines

We use the following performance metrics to measure the evaluation of the algorithms used for predicting the trajectories of vehicles.

- ADE: Average Displacement Error averages Euclidean distances between points of the predicted trajectory and the ground truth that have the same temporal distance from their respective start points.

$$\text{ADE} = \frac{\sum_{n \in N} \sum_{t \in T_{pred}} \| \hat{Y}_t^n - Y_t^n \|_2}{N \times T_{pred}} \quad (6)$$

- FDE: Final Displacement Error measures the distance between final predicted position and the ground truth position at the corresponding time point.

$$\text{FDE} = \frac{\sum_{n \in N} \| \hat{Y}_t^n - Y_t^n \|_2}{N}, t = T_{pred} \quad (7)$$

- RMSE: Root Mean Squared Error evaluates the trajectories in terms of the root of the mean squared error between the prediction results and ground truth.

$$\text{RMSE} = \sqrt{\frac{\sum_{n \in N} \sum_{t \in T_{pred}} (\hat{Y}_t^n - Y_t^n)^2}{N \times T_{pred}}} \quad (8)$$

We compare our approach with the following methods.

- Social LSTM[2]: An LSTM-based network with social pooling of hidden states to predict pedestrian trajectories in crowds.

- CS-LSTM[19]: Adding convolutions to the network in Social LSTM in order to predict trajectories on the highway.

- Social GAN[3]: An LSTM-GAN hybrid network with generative model to predict trajectories for large human crowds.

- Social STGCNN[18]: A Social Spatial-Temporal Graph Convolutional Neural Network, which substitutes the need of aggregation methods by modeling the interactions as a graph.

- STGAT[6]: A Spatial-Temporal Graph Attention network with graph attention mechanism at each time-step to capture the interactions between pedestrians.

### B. Quantitative Results and Analyses

The main results are presented in Table 1. We observe that Spatial-Channel Transformer Network can achieve similar performance results closed to SOTA models. Compared with Social LSTM and CS-LSTM based on LSTM model, our model has an improvement on the precision of trajectories prediction. Furthermore, our model possesses the similar performance with Social GAN. Because the transformer model abandons the recurrent features of LSTM, the cumulative error at the prediction end point is less than previous points. STGAT performs a little better than ours, which proves that Graph Attention Network is promising for modeling interaction between agents. We then further demonstrate this in the following section with visualized results.

Table 2. A result of ablation studies with different range of neighbors. In addition, we compare the performance with/without the SE module in order to evaluate its effect of interaction modeling.

|  |  | Original | No SE module | Original | No SE module | Original | No SE module |
|---|---|---|---|---|---|---|---|
| Vehicle numbers | | 5 | | 10 | | 15 | |
| 1s | ADE/FDE | 0.21/0.41 | 0.22/0.42 | 0.20/0.39 | 0.21/0.40 | 0.20/0.38 | 0.20/0.39 |
| | RMSE | 0.39 | 0.40 | 0.38 | 0.38 | 0.37 | 0.38 |
| 2s | ADE/FDE | 0.52/1.14 | 0.53/1.15 | 0.50/1.10 | 0.51/1.12 | 0.49/1.09 | 0.50/1.10 |
| | RMSE | 0.90 | 0.91 | 0.88 | 0.88 | 0.87 | 0.88 |
| 3s | ADE/FDE | 0.92/2.14 | 0.93/2.15 | 0.88/2.06 | 0.90/2.10 | 0.88/2.07 | 0.89/2.08 |
| | RMSE | 1.54 | 1.56 | 1.51 | 1.52 | 1.51 | 1.51 |
| 4s | ADE/FDE | 1.41/3.37 | 1.42/3.40 | 1.35/3.26 | 1.38/3.30 | 1.36/3.28 | 1.37/3.30 |
| | RMSE | 2.32 | 2.35 | 2.27 | 2.29 | 2.28 | 2.29 |
| 5s | ADE/FDE | 1.97/4.81 | 1.98/4.85 | 1.90/4.66 | 1.93/4.71 | 1.91/4.70 | 1.92/4.73 |
| | RMSE | 3.22 | 3.26 | 3.16 | 3.18 | 3.17 | 3.19 |

*C. Qualitative Results and Analyses*

As shown in Fig. 2, our qualitative results consist of the trajectory visualization in Social GAN model, STGAT model and our spatial-channel model. Our model has the capacity of temporally consistent trajectory prediction. With the given observed trajectory, our model is able to provide proper trajectory results. In addition, our model successfully extracts the social interaction of multiple vehicles to avoid collision. Compared with Social GAN and STGAT, we can see the Spatial-Channel Transformer has a better performance facing a sharp turn situation. It proves that the transformer model has the ability of extracting the spatial-temporal feature of long-term sequence through attention mechanism in the face of variable driving behaviors.

It can be seen from the visualization results that prediction of velocity and change of driving behavior have become important indicators to improve the existing results. Auxiliary information is required for more accurate trajectory prediction, which is not only affected by interaction of neighbors. For the future work, additional information about map and rules of the road, should be added to provide extra information for prediction.

*D. Ablation Studies*

We carry out ablation studies to confirm the effectiveness of channel-wise attention and the selection impact of social neighborhood range.

As shown in Table 2, we arrange the amount of neighborhood agents at 5, 10 and 15. With adding channel-wise attention module, we record the performance of different neighborhood sizes. Generally, channel-wise attention has the capacity of modeling interaction of agents with slight lift of every result of single transformer model. One interesting finding is that the expansion of neighborhood ranges from 5 to 10 will bring out an improvement to prediction results. However, the unlimited expansion of the number of neighbors will not make any improvement further. It can be seen that the result declines slightly when the amounts of agents are increased to 15. It brings invalid information to the training process which is adverse for fitting. Therefore, the number of neighbors is also an important parameter to notice when dealing with the interaction problem between vehicles in trajectory prediction.

V. CONCLUSIONS

In this paper, we propose a Spatial-Channel Transformer Network to predict the trajectories of vehicles on the traffic scenes. We replace common RNN model with Transformer model to process long-term dependency data. With spatial embedding, the spatial-temporal feature of each agent is captured by attention mechanism of the transformer model. To explore social interaction, the channel-wise attention module is jointly used to achieve fully attention functions of trajectory prediction on the NGSIM dataset. It has been proved in the experiments that attention mechanism is powerful to deal with time-series data.

Since trajectory prediction relies on the history data, the uncontrollability of future trajectories still exists. In the future, additional information could be incorporated into the framework to improve our method, such as map data, and other datasets. In addition, the graph structure can also be further combined with existing models to improve the results.